%% file: main.tex
\documentclass[conference]{IEEEtran}
\IEEEoverridecommandlockouts
\usepackage[utf8]{inputenc}
\usepackage[nosort]{cite}
\usepackage{amsmath,amssymb,amsfonts}
\usepackage{graphicx}
\usepackage{textcomp}
\usepackage{xcolor}
\usepackage{algorithm}
\usepackage{multirow}
\usepackage{colortbl}
\usepackage[T1]{fontenc}
\usepackage{textgreek}
\usepackage{caption}
\usepackage{subcaption}
\usepackage{booktabs}
\usepackage{adjustbox}
\usepackage{algpseudocode}

\usepackage{geometry}

\def\BibTeX{{\rm B\kern-.05em{\sc i\kern-.025em b}\kern-.08em
    T\kern-.1667em\lower.7ex\hbox{E}\kern-.125emX}}
\begin{document}

\title{UniCP: A Unified Caching and Pruning Framework for Efficient Video Generation}


\author{
Wenzhang Sun\textsuperscript{1*}, 
Qirui Hou\textsuperscript{2*}, 
Donglin Di\textsuperscript{1},
Jiahui Yang\textsuperscript{2},
Yongjia Ma\textsuperscript{3},\\
Jianxun Cui\textsuperscript{2}
\\
\textsuperscript{1}Li Auto, 
\textsuperscript{2}Harbin Institute of Technology, 
\\
}

\maketitle
  
\renewcommand{\thefootnote}{\fnsymbol{footnote}}
\footnotetext[1]{Equal contribution.}

\begin{abstract}

Diffusion Transformers (DiT) excel in video generation but encounter significant computational challenges due to the quadratic complexity of attention. Notably, attention differences between adjacent diffusion steps follow a U-shaped pattern. Current methods leverage this property by caching attention blocks; however, they still struggle with sudden error spikes and large discrepancies. To address these issues, we propose UniCP—a unified caching and pruning framework for efficient video generation. UniCP optimizes both temporal and spatial dimensions through:
Error-Aware Dynamic Cache Window (EDCW): Dynamically adjusts cache window sizes for different blocks at various timesteps to adapt to abrupt error changes.
PCA-based Slicing (PCAS) and Dynamic Weight Shift (DWS): PCAS prunes redundant attention components, while DWS integrates caching and pruning by enabling dynamic switching between pruned and cached outputs. By adjusting cache windows and pruning redundant components, UniCP enhances computational efficiency and maintains video detail fidelity. Experimental results show that UniCP outperforms existing methods, delivering superior performance and efficiency.

\end{abstract}


\input{section/1-intro}

\input{section/2-relatedwork}

\input{section/3-method}

\input{section/4-experiment}
\input{section/5-conclusion}

\bibliographystyle{IEEEbib}
\bibliography{main}

\end{document}

%% file: section/1-intro.tex
\section{Introduction}
\label{sec:intro}
Diffusion transformers (DiTs) \cite{Peebles2022DiT,yuan2024ditfastattn} have recently become prominent in video generation, often exceeding the output quality of unet-based methods \cite{ho2020denoising,rombach2022high,blattmann2023stable}. However, this advancement requires substantial memory, computational resources, and inference time. Therefore, developing an efficient method for DiT-based video generation is crucial for expanding the scope of generative AI applications.

Unlike traditional unet architectures used in diffusion models \cite{ho2020denoising,song2020denoising}, the DiT employs a distinctive isotropic design that omits encoders, decoders, and skip connections of varying depths.
This causes the existing feature reuse mechanism such as DeepCache \cite{ma2023deepcache} and Faster Diffusion \cite{li2023faster}, may result in the loss of information when applied for DiT. PAB \cite{zhao2024pab} discovered that the attention differences between adjacent diffusion steps follow a U-shaped pattern, and in response developed a pyramidal caching strategy tailored to this observation. $\Delta-$DIT \cite{chen2024delta} discovered that the front blocks primarily handle low-level details, while the back blocks focus more on semantic information, and accordingly designed a two-stage error caching strategy tailored to these insights. DITFastAttn \cite{yuan2024ditfastattn} analyzes redundancy within attention blocks, implementing targeted caching strategies for both the attention outputs and the conditional/unconditional settings. However, these methods typically depend on the U-shaped error curve and manually selected step sizes for caching. As a result, they offer no effective strategies for handling the high-error regions at the ends of the curve or the sudden spikes at its bottom.
\begin{figure}
    \centering
    \includegraphics[width=\linewidth]{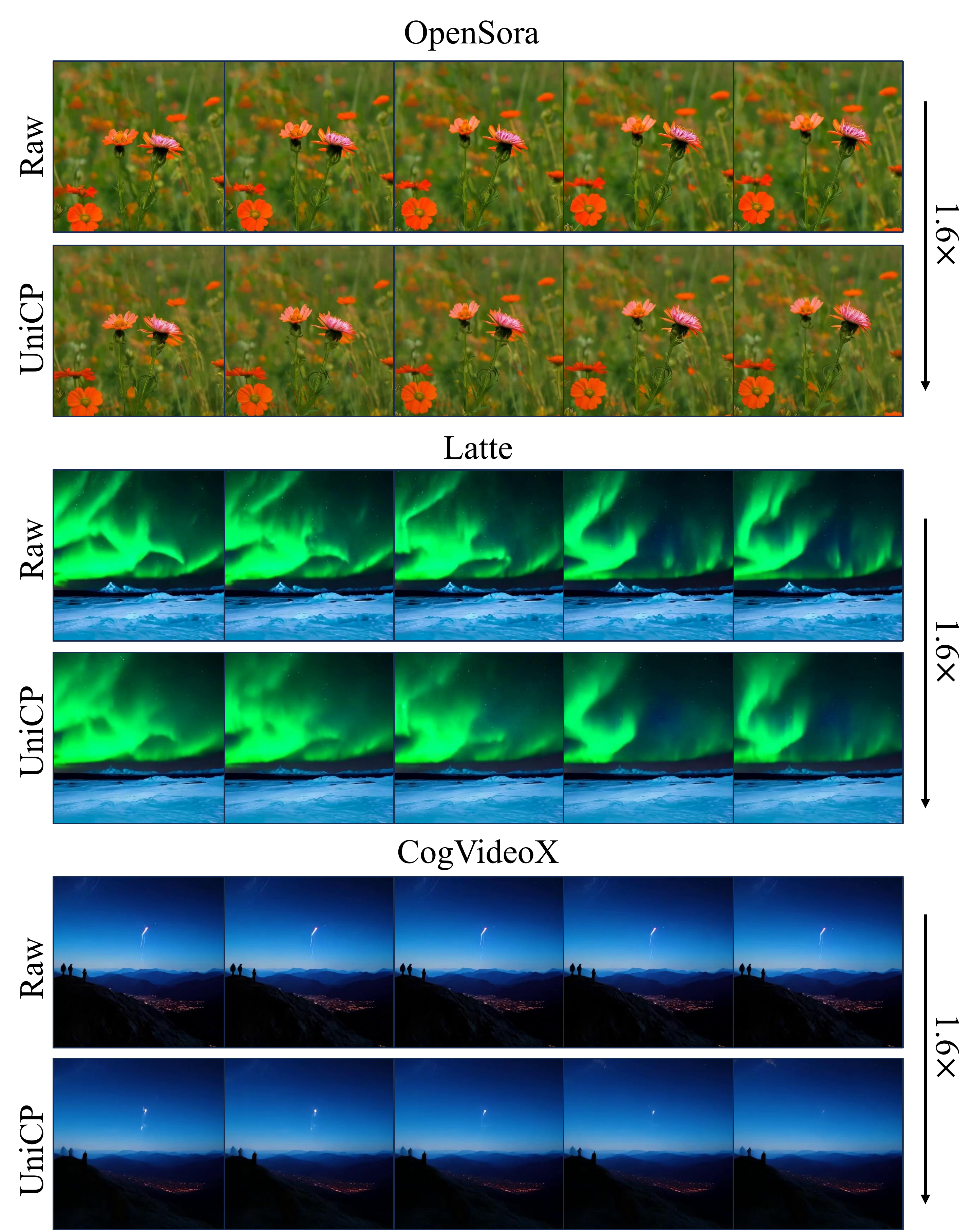}
    \caption{Accelerating video generation methods like OpenSora, Latte, CogVideoX.}
    \label{fig:visual}
    \vspace{-7mm}
\end{figure}
To achieve a more flexible and impactful acceleration solution, we introduce UniCP—an error-aware framework that integrates caching and pruning strategies to accelerate the process across both temporal and spatial dimensions. Specifically: (1) To address sudden error spikes at the bottom of the U-shaped error distribution, UniCP employs an Error-Aware Dynamic Cache Window (EDCW). This mechanism dynamically adjusts caching intervals and strategies based on real-time error feedback. (2) To mitigate large discrepancies at the two ends of the U-shaped error curve, we present a PCA-based Slicing (PCAS) strategy for pruning, further reducing the network’s computational complexity. (3) To unify caching and pruning, we devise a Dynamic Weight Shift (DWS) strategy, seamlessly integrating both approaches across temporal and spatial domains. Our approach delivers up to a 1.6× speedup on a single GPU without compromising video quality. The main contributions of our paper are as follows:

\begin{itemize}
    \item We present UniCP, the first framework to jointly integrate caching and pruning strategies, providing a more flexible and comprehensive approach to accelerating video generation.

    \item An Error-Aware Dynamic Cache Window (EDCW) strategy is proposed to prevent sudden error spikes at the bottom of the U-shaped error distribution.

    \item A PCA-based Slicing (PCAS) strategy is introduced to reduce computational overhead in the attention modules during time steps characterized by large errors that cannot be effectively cached.

    \item A Dynamic Weight Shift (DWS) strategy is proposed to integrate caching and pruning approaches, optimizing the generation process across both spatial and temporal dimensions.
\end{itemize}

%% file: section/2-relatedwork.tex
\section{Related Work}
\label{sec:related_work}
\begin{figure*}[t] 
    \centering
    \includegraphics[width=\textwidth]{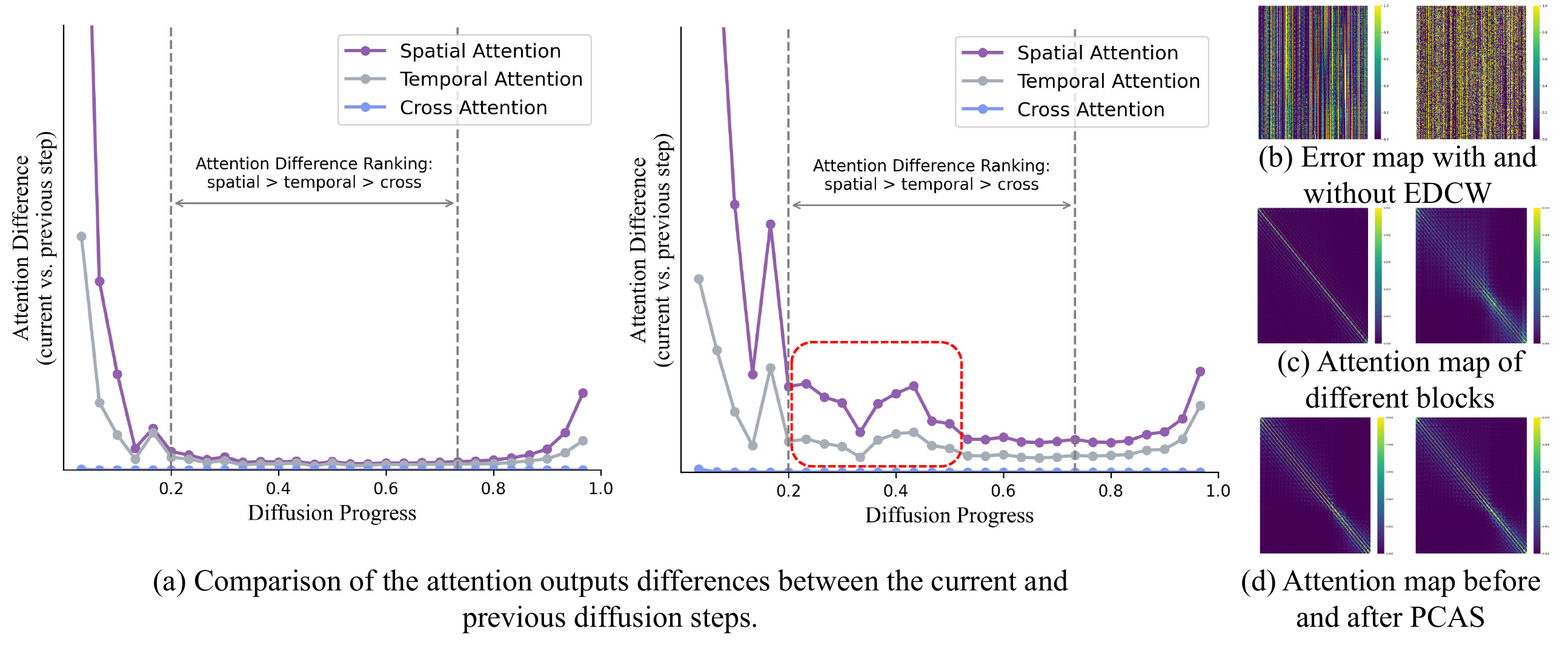} 
    \caption{Visualization of attention differences in Open-Sora. (a) Conventional U-shaped error distribution and sudden error spikes; (b) Error accumulation in regions with sudden spikes: the left side employs the EDCW strategy, while the right side uses manually set cache window sizes; (c) Similarity of attention maps in different blocks; (d) Original attention map and sliced attention map following PCAS.}
    \label{fig:Observations}
    \vspace{-5mm}
\end{figure*}

\subsection{Video Generation}
Video generation focuses on creating realistic videos that are visually appealing and exhibit seamless motion.
The foundational technologies include GAN-based approaches \cite{saito2017temporal,wang2018video}, auto-regressive models \cite{weissenborn2019scaling}, UNet-based diffusion models \cite{blattmann2023stable,ho2022video}, and Transformer-based diffusion models \cite{ma2024latte,hong2022cogvideo,yang2024cogvideox,xu2024vasa}. Among these, diffusion models are widely applied in generating multimodal data, such as video and images \cite{rombach2022high,esser2024scaling}, due to their powerful data distribution modeling capabilities. 
In video generation, Transformer-based diffusion models, specifically those based on DiT, are highly favored for their efficient scalability in an era of increasing computational power.
The computational challenges in Transformer-based frameworks primarily stem from attention mechanisms, where video generation employs three main types: spatial, temporal, and cross attention \cite{ma2024latte,yang2024cogvideox,blattmann2023stable,kong2024hunyuanvideo}. PAB \cite{zhao2024pab} highlights that differences between adjacent diffusion steps are most pronounced in the early and late stages, forming a U-shaped pattern, with significant variations in spatial and temporal attention computations.
This paper specifically addresses the acceleration within the DiT-based video generation framework.

\subsection{Accelerating Diffusion Models}
Video diffusion models have achieved impressive quality in generation, yet their speed is often limited by the sampling mechanisms used during inference. Approaches to accelerate inference can be classified into three main categories: 
(1) Developing enhanced solvers for SDE/ODE equations \cite{lu2022dpm}, which offer limited speed gains and suffer from quality degradation when sampling steps are reduced due to accumulated discretization errors. 
(2) Utilizing diffusion distillation techniques \cite{yin2024onestep}, which apply 2D distillation methods to video generation within a unified diffusion model framework. 
(3) Modifying the architecture of pre-trained models to address computational bottlenecks in the inference process, using techniques such as caching, reuse, and post-training methods like model compression, pruning (\textit{e.g.,} matrix decomposition and dimensionality reduction) \cite{ashkboos2024slicegpt}, and quantization.

Faster Diffusion \cite{li2023faster} caches self-attention early on and then leverages cross-attention for enhancing fidelity in later stages.
PAB \cite{zhao2024pab} eliminates attention computation during the diffusion process by broadcasting attention output in the stable middle phase of diffusion.
$\Delta$-DiT \cite{chen2024delta} leverages the correlation between DiT blocks and image generation by caching backend blocks in early sampling stages and frontend blocks in later stages to achieve faster generation.
Unlike these methods, we focus on accelerating the computation of temporal and spatial attention by utilizing caching and post-training techniques (in our paper, \textit{i.e.,} PCA dimensionality reduction), which are commonly used in the Natural Language Processing field \cite{ashkboos2024slicegpt}.

%% file: section/3-method.tex
\section{Method}
\label{sec:method}

\begin{figure}[t] 
    \centering
    \includegraphics[width=0.48\textwidth]{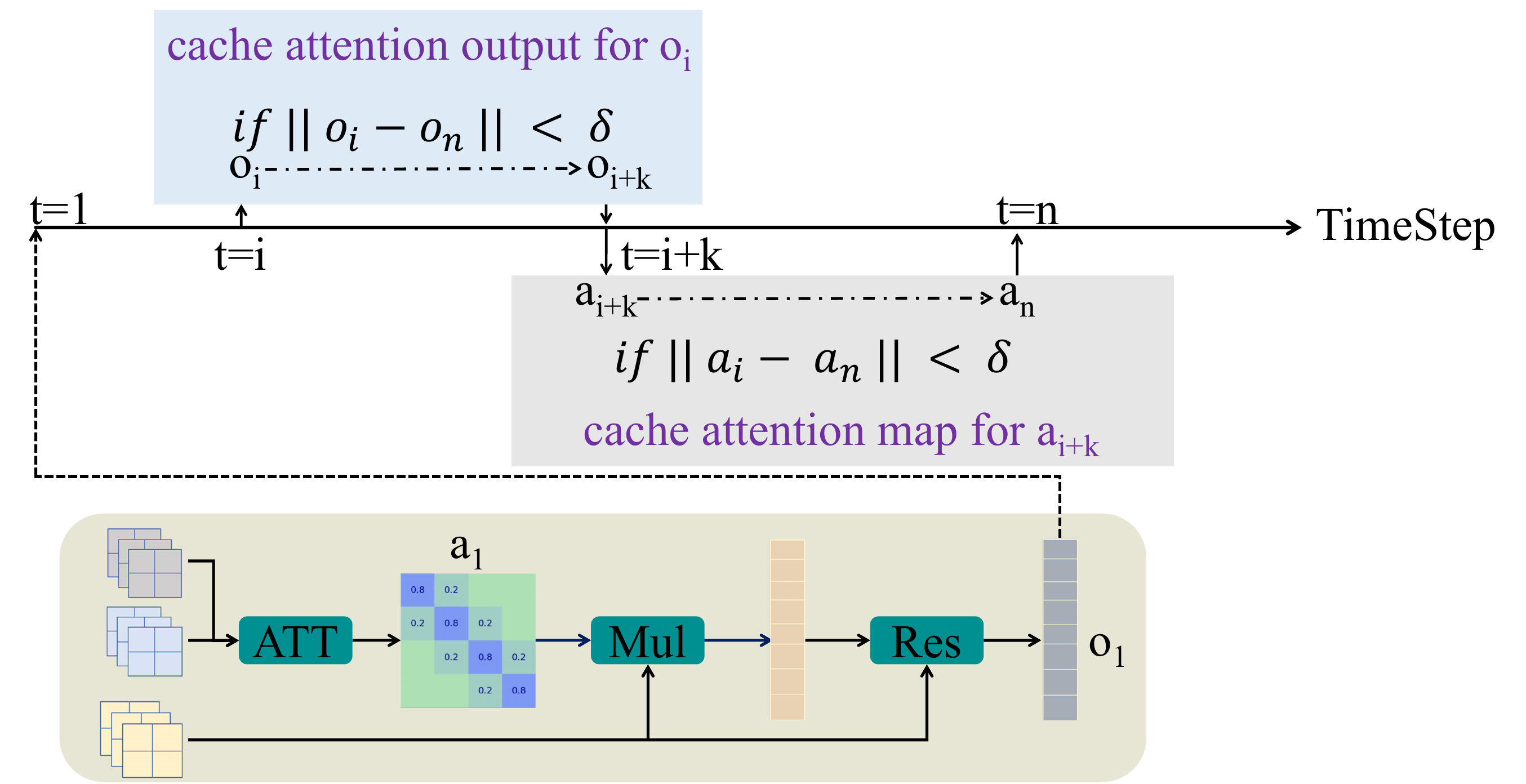} 
    \caption{Visualization of the cache routine in EDCW. EDCW dynamically adjusts the cache window size and caching strategy based on the error threshold.}
    \label{fig:cache}
    \vspace{-3mm}
\end{figure}
\begin{figure}[t] 
    \centering
    \includegraphics[width=0.48\textwidth]{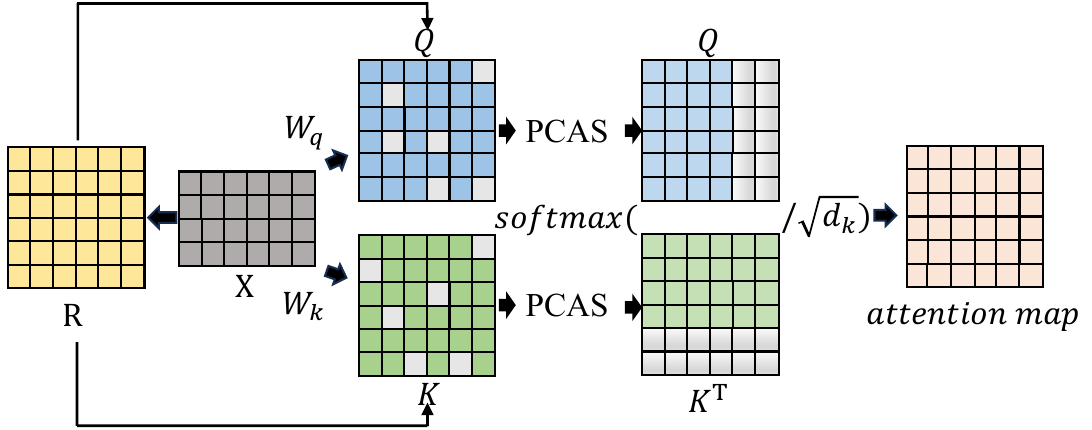} 
    \caption{Visualization of the PCAS. PCAS reduces the computational cost of the attention mechanism by pruning redundant dimensions in the query and key matrices.}
    \label{fig:slice}
    \vspace{-4mm}
\end{figure}
\begin{figure}[t]
    \centering
    \includegraphics[width=\columnwidth]{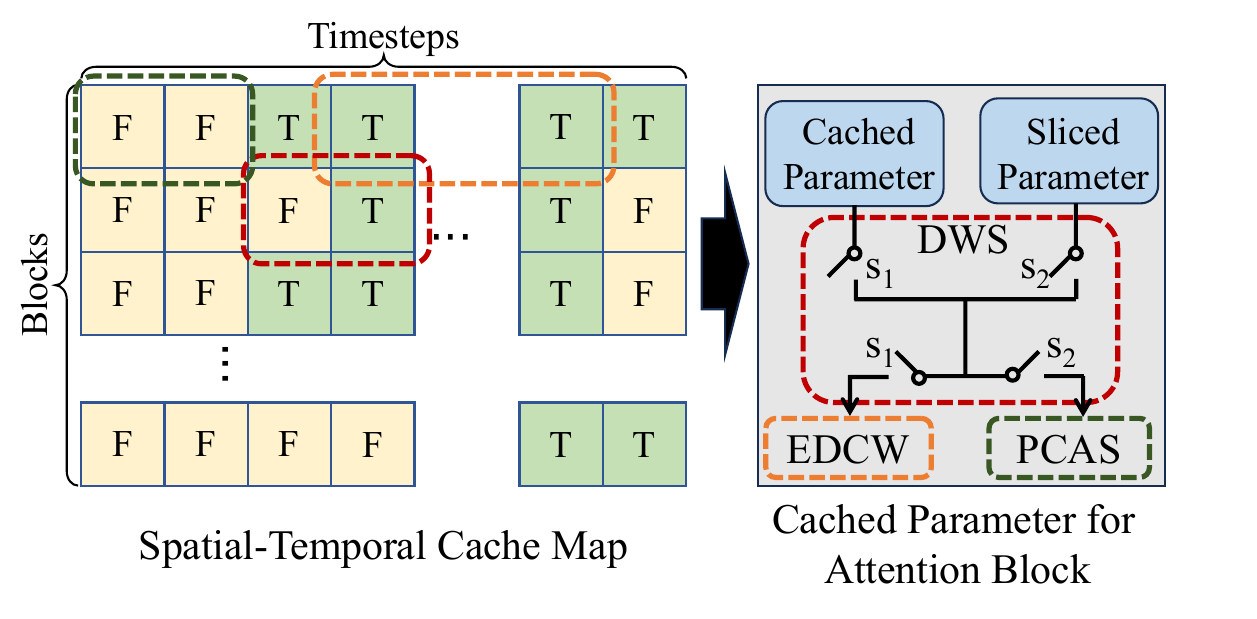}
    \caption{After acquiring the spatial-temporal cache map, the DWS strategy enables dynamic switching between caching and pruning strategies, allowing both processes to operate within a unified framework.}
    \label{fig:Cache map}
\vspace{-4mm}
\end{figure}
As illustrated in Fig.~\ref{fig:Observations}, even at the bottom of the U-shaped error curve, sudden error spikes can still occur. Manually setting the caching interval results in unstable outcomes. Moreover, significant errors on both sides of the U-shaped error distribution lead to poor performance when employing a caching strategy. To address these challenges, we propose three targeted optimization strategies. In Section \ref{chap:EDCW}, we introduce Error-Aware Dynamic Cache Windows, which dynamically adjust the caching window in response to observed errors. In Section \ref{chap:PCAS}, we present PCA-based Slicing to further reduce computation during time steps that cannot be cached due to error distribution. In Section \ref{chap:DWS}, we describe a Dynamic Weight Shift strategy that integrates pruning and caching strategies into a unified framework.
\subsection{Error-Aware Dynamic Cache Window (EDCW)}
\label{chap:EDCW}
As shown in Fig.~\ref{fig:Observations}, the error distribution within each attention block during inference does not strictly form a U-shaped pattern. Instead, there are larger errors at both ends and sudden spikes at the bottom. We contend that the caching interval should be determined by the error itself. Given a certain attention block $i$ under timestep $j$, the cache step $t_{i,j} $ can be defined as:
\begin{equation}
    t_{i,j}=\omega(\delta , o_{i}, o_{i+k}, a_{i}, a_{i+k})
\end{equation} 
Where \(\delta\) is the user-defined error threshold, and \(k=K,K-1,\ldots,1\) denotes the size of the dynamic search window. The parameter \(\omega\) specifies a dynamic caching strategy: we begin by measuring the error between the attention outputs \((o_i)\) and \((o_j)\). If the computed error fails to meet the threshold \(\delta\), an alternative caching approach is then applied to the attention map \((a_i)\). The first strategy offers greater computational savings but allows for higher error, whereas the second approach preserves more accuracy at the cost of reduced computational gains. By employing predefined error thresholds, EDCW can dynamically adjust both the caching window and the chosen caching strategy. 

\subsection{PCA-based Slicing (PCAS)}
\label{chap:PCAS}
At the far ends of the U-shaped error distribution — encompassing roughly 30\% of all steps — attention block outputs diverge significantly. In such situations, attempting to cache results may actually degrade the method’s performance. To mitigate this, we introduce a pruning approach tailored for those uncachable blocks, focusing on the query and key transformations, as well as the corresponding linear layer parameters, to further alleviate computational burden.

Principal component analysis (PCA) commonly aims to transform an original data matrix \(\mathbf{X} \in \mathbb{R}^{m \times m}\) into a compact representation \(\mathbf{\overline{Z}} \in \mathbb{R}^{m \times n}\) (where \(n < m\)) and a reconstructed approximation \(\overline{\mathbf{X}} \in \mathbb{R}^{m \times m}\). The core operation of an attention map involves multiplying the query and the key matrices, expressed as: $\text{softmax}(\mathbf{Q}\mathbf{K}^{\top})$. Given an attention input $\mathbf{X}$ and the corresponding weight for the query and key, it can be length defined as: $\text{softmax}((\mathbf{XW_{q}})(\mathbf{XW_{k}}^{\top}))$. Letting \(\mathbf{R} \in \mathbb{R}^{m \times m}\) represent an eigenvector matrix of the attention input \(\mathbf{X}^{\top}\mathbf{X}\), the compressed expression for query and key can be formulated as:
\begin{equation}
        \mathbf{Z_{q}} = \mathbf{(XW_{q})RD}, \quad \quad \overline{\mathbf{Q}} = \mathbf{Z_{q} D}^{\top}\mathbf{R}^{\top}.
\end{equation}
\begin{equation}
        \mathbf{Z_{k}} = \mathbf{(XW_{k})RD}, \quad \quad \overline{\mathbf{K}} = \mathbf{Z_{k} D}^{\top}\mathbf{R}^{\top}.
\end{equation}
Here, \(\mathbf{D} \in \mathbb{R}^{m \times n}\) is a selection matrix derived from the identity matrix, retaining only \(n\) thin columns to reduce dimensionality while preserving critical structure. This reconstruction is \(L_2\)-optimal in the sense that the chosen linear mapping $\mathbf{RD}$ minimizes \(\|\mathbf{X} - \overline{\mathbf{X}}\|\).

\begin{algorithm}
\caption{Detailed Caching and Pruning Strategy Given an Attention Block under Certain Timestep}
\begin{algorithmic}
\Require $X, \text{δ}$ \Comment{Attention Input, Error Threshold}
\Require $K, c$ \Comment{Search Window, Cache Status}
\Require $a, o$ \Comment{Attention Map, Attention Output}
\Require $i$ \Comment{Current Step}

\State Initialize Cache Status $c=F$


\If{$c=F:$}
    \For{$k=K$ to $1$:}
\State \textbf{if }{$\begin{Vmatrix}o_{i} - o_{i+k}\end{Vmatrix} _{2} \le \delta_{i+k}:$}
        cache attention output; $c=T$; Break;
    \EndFor
\For{$k=K$ to $1$:}
\State \textbf{if }{$\begin{Vmatrix}a_{i} - a_{i+k}\end{Vmatrix} _{2} \le \delta_{i+k}:$}
        cache attention map; $c=T$; Break;
    \EndFor
\State Apply PCA-based Slicing
\State Update c as processed; Break;
\EndIf

\end{algorithmic}
\end{algorithm}

\subsection{Dynamic Weight Shift (DWS)}
\label{chap:DWS}

EDCW and PCAS enhance the network from spatial (individual attention block processing) and temporal (caching across denoising steps) dimensions, but applying both simultaneously can cause interference. To unify them, we propose a Dynamic Weight Shift (DWS) strategy. Guided by a cache map Fig.~\ref{fig:Cache map}, DWS identifies uncachable blocks, prunes them selectively, and stores both pre-pruning and post-pruning weights, allowing for a dynamic integration of pruning and caching. As shown in Algorithm 1, we initially skip partitioning to preserve the original output. Then, we gradually increase the attention head partition size until the loss approaches the threshold curve, recording a suitable dimension \(k\) that remains below this threshold. Once the adaptive algorithm finishes, we compile all recorded \(k\) values for that block and select the smallest one as the final pruning dimension.



%% file: section/4-experiment.tex
\section{Experiment}\label{sec:exp}

\begin{table*}[ht]
\centering
\caption{Performance of UniCP across Different Video Generation Models. We assessed UniCP under varying error thresholds. The top three results are distinguished by color: blue indicates the first rank, red the second, and green the third.}
\label{table:Comparison of efficiency and visual quality}

    \begin{tabular}{ccccccccc}
        \toprule
        model & method & MACs (P) $\downarrow$ & Speedup $\uparrow$ & Latency (s) $\downarrow$ & VBench$\uparrow$ & LPIPS$\downarrow$ & SSIM $\uparrow$ & PSNR $\uparrow$ \\ 
        \midrule
        \multirow{4}{*}{Open-Sora} 
        & origin & 5.59 & 1$\times$ & 54.38& 78.69\% & - & - & - \\
        & PAB \cite{zhao2024pab} & 4.81 & 1.34$\times$ & 40.59 & \textcolor{blue}{78.21\%} & 0.1020 & 0.8821 & 26.43 \\
        & FasterCache \cite{lv2024fastercache} &  \textcolor{red}{4.13} & \textcolor{red}{1.57$\times$} & \textcolor{red}{34.64} & 77.69\% & 0.0937 & 0.8830 & 26.57 \\
        \hline
        &UniCP (E1) & 5.23 & 1.11$\times$ & 49.07 & \textcolor{red}{78.17\%} & \textcolor{blue}{0.0847} & \textcolor{blue}{0.9017} & \textcolor{blue}{27.15} \\
        &UniCP (E2) &5.05&1.16$\times$&46.72&\textcolor{cyan}{78.03\%}&\textcolor{red}{0.0857}&\textcolor{red}{0.8970}&\textcolor{red}{26.99}\\
        &UniCP (E3) &4.82&1.29$\times$&42.29&77.92\%&\textcolor{cyan}{0.0879}&\textcolor{cyan}{0.8917}&\textcolor{cyan}{26.57}\\
        &UniCP (E4) &\textcolor{cyan}{4.47}&\textcolor{cyan}{1.42} &\textcolor{cyan}{38.33}&77.41\%&0.0893&0.8871&26.43\\
        &UniCP (E5) & \textcolor{blue}{4.09} & \textcolor{blue}{1.59$\times$} & \textcolor{blue}{34.20} & 77.34\% & 0.0892 & 0.8874 & 26.37 \\
        \midrule
        \multirow{4}{*}{Latte}
        & origin &3.05 & 1$\times$ & 28.71 & 77.23\% & - & - & - \\
        & PAB \cite{zhao2024pab} & 2.24 & 1.32$\times$ & 21.76 & \textcolor{cyan}{77.01\%} & 0.2837 & 0.7152 & 19.20 \\ 
        & FasterCache \cite{lv2024fastercache} & \textcolor{red}{1.97} & \textcolor{red}{1.59$\times$} & \textcolor{red}{18.06} & 76.92\% & 0.0952 & 0.8548 & 23.75 \\ 
        \hline
        &UniCP (E1) & 2.79 & 1.10$\times$ & 26.03 & \textcolor{blue}{77.13\%} & \textcolor{blue}{0.0827} & \textcolor{blue}{0.9019} & \textcolor{blue}{25.34} \\
        &UniCP (E2) &2.67&1.19$\times$&24.03&\textcolor{red}{77.05\%} &\textcolor{red}{0.0878}&\textcolor{red}{0.8872}&\textcolor{red}{24.89}\\
        &UniCP (E3) &2.45&1.29$\times$&22.11&76.96\%&\textcolor{cyan}{0.0912}&\textcolor{cyan}{0.8734}&\textcolor{cyan}{24.46}\\
        &UniCP (E4) &\textcolor{cyan}{2.21}&\textcolor{cyan}{1.44$\times$}&\textcolor{cyan}{19.87}&76.89\%&0.0934&0.8541&24.05\\
        &UniCP (E5) & \textcolor{blue}{1.96} & \textcolor{blue}{1.61$\times$} & \textcolor{blue}{17.83} & 76.82\% & 0.0978 & 0.8471 & 23.65 \\
        \midrule
        \multirow{4}{*}{CogVideoX}
        & origin & 6.03 & 1$\times$ & 85.74 & 81.18\% & - & - & - \\ 
        & PAB \cite{zhao2024pab} & 4.45 & 1.32$\times$ & 65.06 & \textcolor{red}{79.85\%} & 0.0872 & \textcolor{cyan}{0.9463} & 28,51 \\
        & FasterCache \cite{lv2024fastercache} & \textcolor{red}{3.71} & \textcolor{red}{1.60$\times$} & \textcolor{red}{53.61} & 78.34\% & 0.0850 & \textcolor{blue}{0.9572} & \textcolor{red}{28.66} \\
        \hline
        &UniCP (E1) & 5.35 & 1.12$\times$ & 76.91&\textcolor{blue}{79.97\%} & \textcolor{blue}{0.0835} & \textcolor{red}{0.9479} & \textcolor{blue}{28.87} \\
        &UniCP (E2) &4.96&1.21$\times$&70.88&\textcolor{cyan}{79.25\%}&\textcolor{red}{0.0840}&0.9411&28.61\\
        &UniCP (E3) &4.54&1.33$\times$&64.51&78.72\%&\textcolor{cyan}{0.0845}&0.9385&\textcolor{cyan}{28.63}\\
        &UniCP (E4) &\textcolor{cyan}{4.11}&\textcolor{cyan}{1.49$\times$}&\textcolor{cyan}{57.46}&78.39\%&0.0860&0.9299&28.44\\
        &UniCP (E5) &\textcolor{blue}{3.61}&\textcolor{blue}{1.64$\times$}&\textcolor{blue}{50.37}&77.72\%&0.0866&0.9237&28.37\\
        
        \bottomrule
    \end{tabular}

\end{table*}
In this section, we describe experimental setup and present the results and key findings from experiments.
\subsection{Experimental Setup}
Our method is integrated into state-of-the-art DIT-based video generation models, including OpenSora 1.2, Latte, and CogVideoX These models serve as the foundation for our experiments, allowing us to assess the effectiveness of our proposed approach. For baseline comparisons, we employ PAB \cite{zhao2024pab} and FasterCache \cite{lv2024fastercache}, both of which are based on caching frameworks. Additionally, we utilize the prompts provided by the VBench as our evaluation dataset to comprehensively evaluate performance. All experiments were conducted on NVIDIA A800 80GB GPUs.

\subsection{Evaluation Metrics} To assess the visual quality of generated videos, we utilize several metrics, including VBench\cite{huang2024vbench}, LPIPS\cite{zhang2018unreasonable}, SSIM\cite{wang2004image}, and PSNR\cite{korhonen2012peak}. VBench provides a standardized benchmarking framework for comparing various algorithms. LPIPS measures perceptual similarity by computing distances in the image feature space using pretrained convolutional neural networks. SSIM evaluates image similarity by considering luminance, contrast, and structural information. PSNR  quantifies video quality by measuring the error between video sequences, offering a precise indication of their differences. Additionally, to evaluate latency and computational complexity, we use Latency (inference time) and Multiply-Accumulate Operations (MACs). These metrics are essential for quantifying the computational cost during inference process and are robust indicators of acceleration method effectiveness.

\begin{figure}
    \centering
    \includegraphics[width=\linewidth]{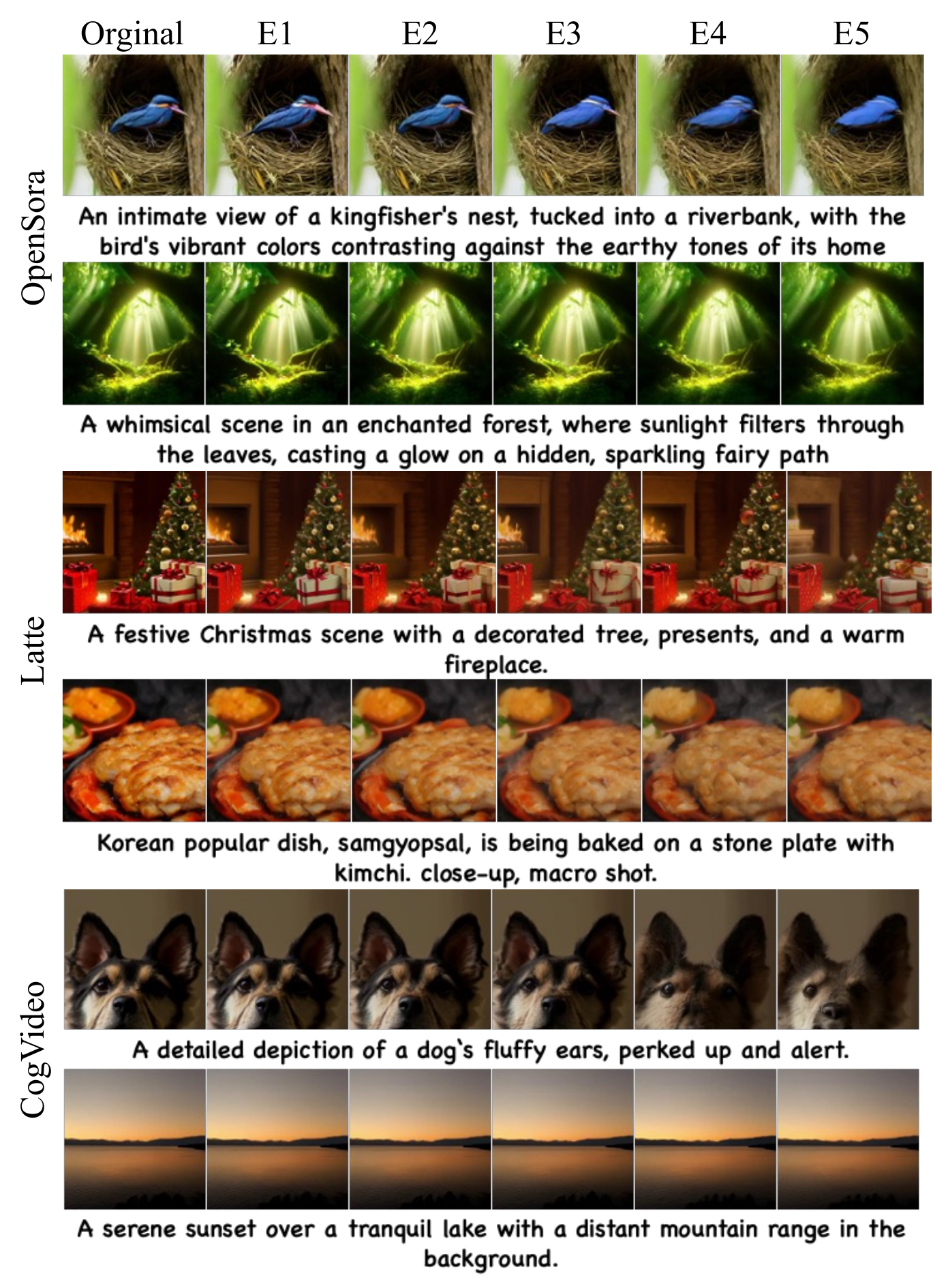}
    \caption{Video generation samples under various error thresholds. UniCP demonstrates stable performance across various error thresholds, with only minimal quality degradation.}
    \label{fig:maxinexp}
    \vspace{-5mm}
\end{figure}
\subsection{Quantitative Experiments}
Quantitative experiments with state-of-the-art methods are presented in Table \ref{table:Comparison of efficiency and visual quality}. We synthesize videos using prompts provided by VBench and employ these synthesized videos to compute the VBench metrics. Additionally, we calculate LPIPS, SSIM, and PSNR using videos generated by the original models. We denote these threshold settings as E1 (δ=0.025), E2 (δ=0.05), E3 (δ=0.75), E4 (δ=0.125), E5 (δ=0.175). The results indicate that UniCP maintains stable performance across various error thresholds. As the threshold increases, it significantly reduces computational complexity and latency while largely preserving video quality.



\subsection{Qualitative Experiments} Consistent with the experimental setup described earlier, we visualize the video results generated under different error thresholds (E1, E2, \ldots, E5). The generated images are presented in Fig.~\ref{fig:maxinexp}. In these visual comparisons, our method demonstrates a remarkable ability to maintain video quality, particularly in terms of color accuracy and detail preservation.

\subsection{In-depth Analysis}
To investigate the acceleration potential and characteristics of our strategy, we conducted extensive ablation experiments. In the following experiments, we deployed Open-Sora 1.2 as the base model and used a single Nvidia A800 GPU to generate 49-frames videos.

\begin{figure}[t]
    \centering
    \includegraphics[width=\linewidth]{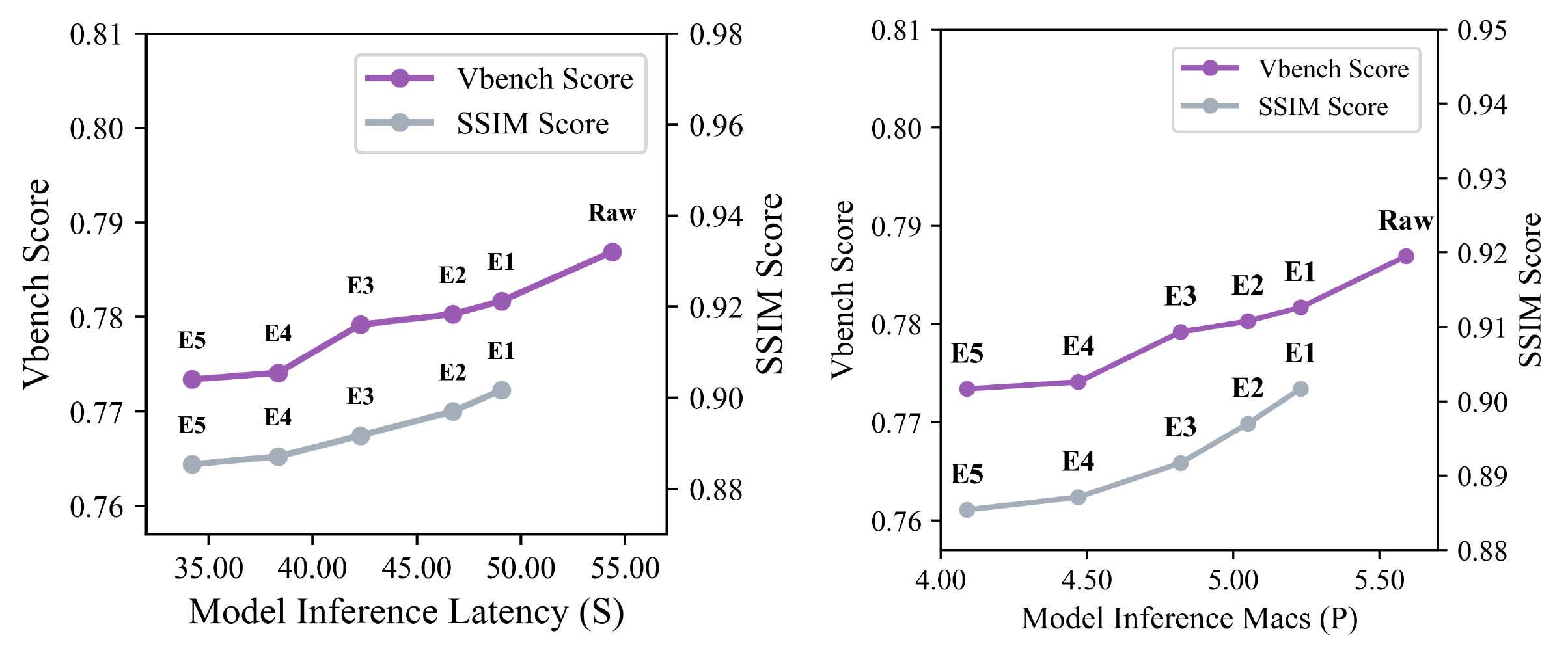}
    \caption{Visualization of generated video quality, latency, and computational complexity under different error thresholds.}
    \label{fig:flops_comparison}
    \vspace{-4mm}
\end{figure}
\begin{table}[t]
\caption{Quantitative analysis of different caching strategies.}
\label{table:Comparison of different strategies}
\centering
\resizebox{\columnwidth}{!}{
\begin{tabular}{lccc}
\toprule
Strategy         & Latency (s) & $\Delta$ & VBench (\%) $\uparrow$ \\ \hline
attention output & 49.19       & 5.19     & \textbf{78.23}                  \\
attention map    & 49.27       & 5.11     & 78.26                  \\
dynamic select   & \textbf{49.07}       & \textbf{5.31}     & 78.17                  \\ \bottomrule
\end{tabular}
}
\vspace{-4mm}
\end{table}
\noindent{\textbf{Error Thereshold Analysis.}} We evaluated the computational complexity, latency, and video quality of models compressed with UniCP on OpenSora across different error thresholds (Fig. \ref{fig:flops_comparison}). Results demonstrate that increasing the error threshold leads to a significant reduction in both computational complexity and latency, while the generated video quality remains largely stable, exhibiting only minor decreases.
\begin{figure}[t]
    \centering
    \includegraphics[width=0.98\linewidth]{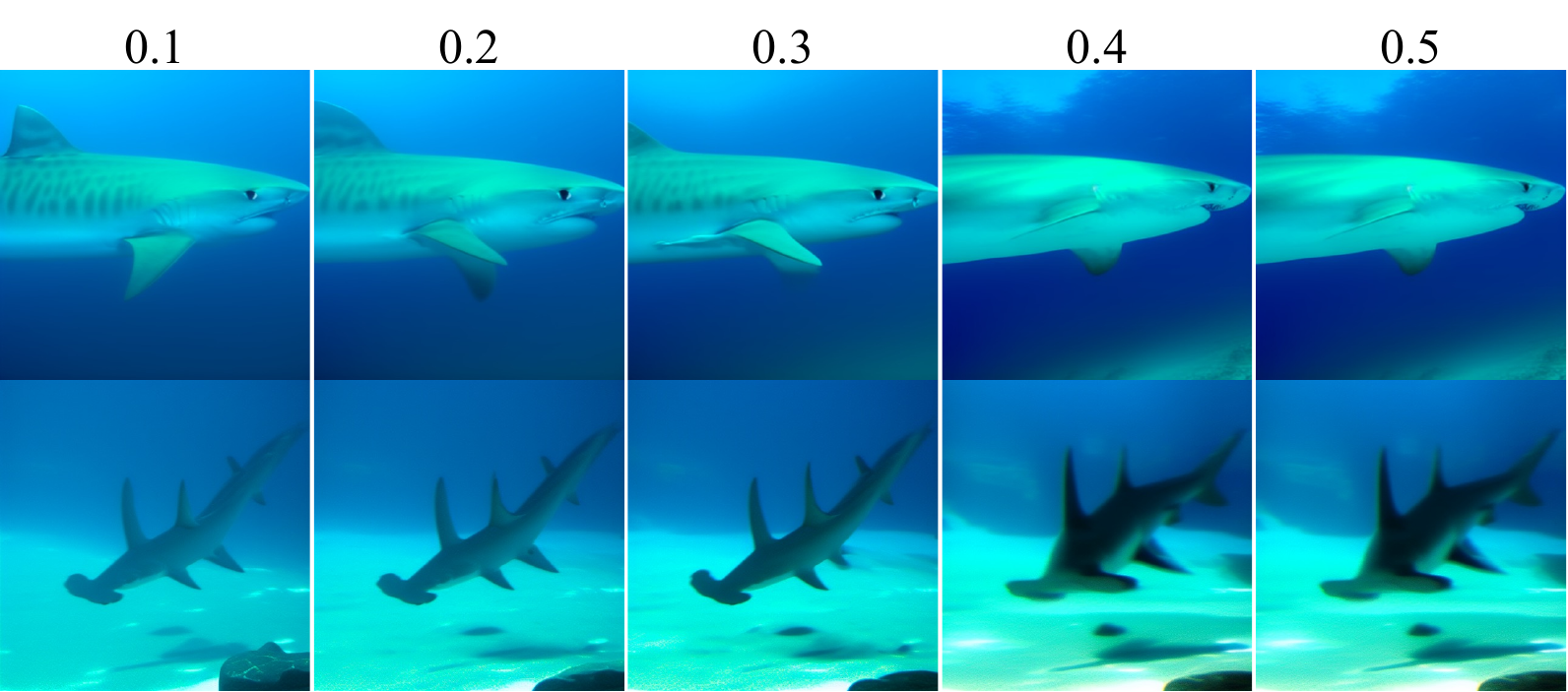}
    \caption{Visual results across various slice ratios.}
    \label{fig:slice1}
    \vspace{-4mm}
\end{figure}

\noindent{\textbf{Caching Strategy Analysis.}} Caching entire blocks typically induces significant errors. To address this, we developed dynamic caching strategies for the attention output and attention map. TABLE \ref{table:Comparison of different strategies} presents the quantitative compression results on OpenSora with an error threshold of 0.025. The results show that caching the attention map achieves greater computational savings but leads to more video quality degradation. In contrast, our proposed strategies reduce computational overhead while maintaining video quality.


\noindent{\textbf{Slice Ratio Analyse.}} Fig. \ref{fig:slice1} illustrates the performance of the PCAS strategy across different partition ratios. High image quality is maintained when the partition ratio is below 0.4. In this work, we dynamically adjust the partition ratio within the range of 0.1 to 0.4, adhering to the defined error threshold.

%% file: section/5-conclusion.tex
\section{Conclusion}
\label{sec:conclusion}
We presents UniCP, a novel model acceleration method that unifies caching and pruning strategies within a single framework. To address the diverse error distributions observed across different blocks during the network denoising process, we introduce an Error-Aware Dynamic Cache Window, which dynamically adjusts both the caching step size and strategy. Furthermore, to eliminate redundant computations in areas with substantial error variations, we employ PCA-based Slicing. Lastly, the Dynamic Weight Shift strategy seamlessly integrates caching and pruning methodologies. Applied to various video generation models, UniCP significantly improves runtime efficiency while preserving video quality.